\documentclass[wcp]{jmlr}

\usepackage{longtable}

\usepackage{booktabs}
\usepackage{amsfonts}
\usepackage{graphicx}
\usepackage{amsmath}
\usepackage{amssymb}
\usepackage{adjustbox}
\usepackage{multirow}
\usepackage{subfloat} 
\usepackage{bbm}
\usepackage{float}
\usepackage{changes}
\usepackage{threeparttable}
\usepackage{bm}
\usepackage{booktabs,makecell,multirow}
\makeatletter
\let\Ginclude@graphics\@org@Ginclude@graphics 
\makeatother

\jmlrvolume{157}
\jmlryear{2021}
\jmlrworkshop{ACML 2021}

\title[Pyramid Correlation based Deep Hough Voting \\ for Visual Object Tracking]{Pyramid Correlation based Deep Hough Voting \\ for Visual Object Tracking}

  \author{\Name{Ying Wang} \Email{wangying7275@gmail.com}\\
  \Name{Tingfa Xu$^{\ast}$} \Email{ciom$\_$xtf1@bit.edu.cn}\\
  \Name{Jianan Li$^{\ast}$} \Email{lijianan15@gmail.com}\\
  \Name{Shenwang Jiang} \Email{jiangwenj02@gmail.com}\\
  \Name{Junjie Chen} \Email{3120190516@bit.edu.cn}\\
  \addr Beijing Institute of Technology, Beijing, China
  \thanks{$^\ast$Tingfa Xu and Jianan Li are the corresponding authors.}
 }

\editors{Vineeth N Balasubramanian and Ivor Tsang}

\begin{document}

\maketitle

\begin{abstract}
Most of the existing Siamese-based trackers treat tracking problem as a parallel task of classification and regression.
However, some studies show that the sibling head structure could lead to suboptimal solutions during the network training. Through experiments we find that, without regression, the performance could be equally promising as long as we delicately design the network to suit the training objective. We introduce a novel voting-based classification-only tracking algorithm named Pyramid Correlation based Deep Hough Voting (short for PCDHV), to jointly locate the top-left and bottom-right corners of the target. Specifically we innovatively construct a Pyramid Correlation module to equip the embedded feature with fine-grained local structures and global spatial contexts; The elaborately designed Deep Hough Voting module further take over, integrating long-range dependencies of pixels to perceive corners; In addition, the prevalent discretization gap is simply yet effectively alleviated by increasing the spatial resolution of the feature maps while exploiting channel-space relationships. The algorithm is general, robust and simple. We demonstrate the effectiveness of the module through a series of ablation experiments. Without bells and whistles, our tracker achieves better or comparable performance to the SOTA algorithms on three challenging benchmarks (TrackingNet, GOT-10k and LaSOT) while running at a real-time speed of 80 FPS. Codes and models will be released.
\end{abstract}


\section{Introduction}
\label{sec:intro}

Visual object tracking is a fundamental task in computer vision, whose purpose is to predict the position of a target in subsequent frames given its precise state in the initial frame. It has been widely used in applications such as surveillance, robotics, autonomous driving~\cite{7637024,li2013survey,kiani2017learning}. Despite remarkable progress made in recent decades, challenges such as large occlusion, severe deformation and similar object interference still stand to be overcome~\cite{7001050,6784124}.

\begin{figure*}
\begin{center}
\includegraphics[width=1\linewidth]{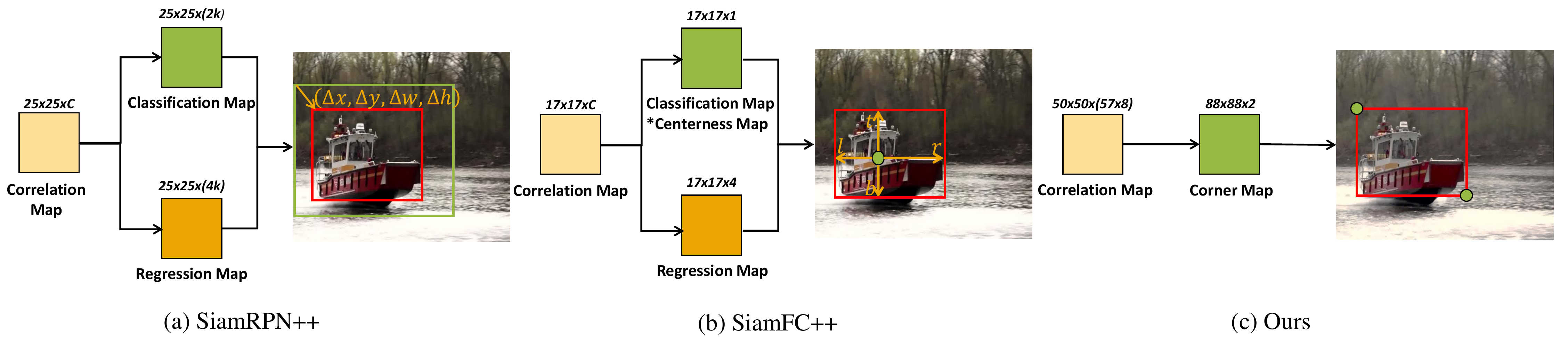}
\end{center}
   \caption{
    Structure comparison of (a) SiamRPN++, (b) SiamFC++ and (c) our approach. Symbols in red, orange and green represent the classification branch, the regression branch and the final predicted target box, respectively. Contrary to previous algorithms, we discard the sibling heads model and turn to locate the target sorely by the classification branch, separating corner points from background through a voting mechanism.
}
\label{fig:evolution}
\end{figure*}

Siamese-based algorithms contribute significantly to the field of visual tracking. 
It treats the tracking task as a target matching problem by learning the general similarity map between the target template and the search region. 
Recent years have witnessed a greater tendency for many Siamese-based algorithms ~\cite{Li_2018_CVPR,Li_2019_CVPR,Danelljan_2019_CVPR,Voigtlaender_2020_CVPR,10.1007/978-3-030-58589-1_46,yan2021learning} to treat tracking problem as a combination of a parallel classification and regression. 
To be specific, anchor-based algorithms SiamRPN~\cite{Li_2018_CVPR}, SiamRPN++~\cite{Li_2019_CVPR}, DaSiam~\cite{10.1007/978-3-030-01240-3_7} introduce the region proposal extraction subnetwork (RPN)~\cite{7485869} and use a classification branch for proposal selection, further regressing the four offsets between the anchor and the corresponding groundtruth (as illustrate in Fig.\ref{fig:evolution} (a)). While anchor-free tracking algorithms SiamFC++~\cite{Xu_Wang_Li_Yuan_Yu_2020}, SiamCAR~\cite{Guo_2020_CVPR} classifies all positions within the object bounding box as positive with the help of centerness branch and regresses the four distances between the center-point and the the object boundaries (as illustrate in Fig.\ref{fig:evolution} (b)).
Although these approaches obtain balanced accuracy and speed, it has been proposed that the tracking performance is prone to fall into suboptimal due to the essential misalignment of the two branches and that substantially the classification branches contribute more~\cite{Song_2020_CVPR,Cheng_2018_ECCV,yan2021learning}.
Therefore, in our work, we design a voting-based classification-only tracking algorithm (short for PCDHV), solely using a classification branch to generate corner voting map, jointly locating the top-left and bottom-right corners of the target (as illustrate in Fig.\ref{fig:evolution} (c)).
The whole structure consists of three modules: Siamese-based feature extraction, Pyramid Correlation and Deep Hough Voting.

Concretely, the traditional Siamese structure is adopted first for feature extraction of target template and the search region, whose outputs are then embedded through cross-correlation to learn the similarity. Most of the existing Siamese-based algorithms employ naïve-correlation~\cite{SiamFC,Li_2018_CVPR} and depth-wise correlation~\cite{Li_2019_CVPR,Xu_Wang_Li_Yuan_Yu_2020}, taking the entire template feature as a convolution kernel.
However, in the process of tracking, the target may suffer large appearance variance, thus matching a fixed kernel feature as a whole with the vastly changing search features may drastically degrade the quality of correlation map.
Considering the corner locating objective which requires the correlation feature rich in fine-grained local structure, we argue that the pixel-wise correlation is more suitable.
Innovatively, we propose Pyramid Correlation to extract corner-favoriable fusion feature through a series of blocks: spatial feature selection, pyramid feature pooling and group pixel-level correlation. These blocks simultaneously equip the correlation feature with global spatial contexts, making the tracker robust to deformation.

Besides, corner locating also requires long-range dependencies of pixels. To this end, we elaborately design a Deep Hough Voting module to further take over, which consists of three blocks of vote generation, vote refinement and vote aggregation, precisely produce feature maps representing the location probabilities of the top-left and bottom-right corners.
The vote generation block first applies several convolutional layers to adjust feature into appropriate shape, expanding the receptive field at the same time.
Then vote refinement block, namely a modified position-aware non-local block, mines the dense contextual information and pairing two corner feature more closely. Grid channel and Pixel Shuffle mechanism~\cite{7780576} are deployed to obtain location information and expand feature size respectively, further exploiting the channel-spatial relationships.
Finally, the voting module proposed in HoughNet~\cite{10.1007/978-3-030-58595-2_25} takes responsibility of the voting aggregation block to capture long-range dependencies.
The three blocks work together to produce high-quality corner heatmaps, with the peak position of each map representing the location of the predicted corner point.

Inevitably, the discretization gap caused by the stride of the network has a dramatic negative impact on the tracking accuracy.
While many superior algorithms resort to introducing regression branch to bridge this gap, we instead choose to compensate by inserting up-sampling operations in the structure. As a result, our PCDHV can localize the bounding box in a precise way, as can be seen intuitively from the experimental results below.

We evaluate our PCDHV on three challenging large-scale benchmarks, including GOT-10k~\cite{huang2019got}, TrackingNet~\cite{2018TrackingNet} and LaSOT~\cite{fan2019lasot}.
Without bells and whistles, our tracker can achieve better or comparable performance to the state-of-the-art trackers. 
The network is general, robust yet simple, requires no tedious parameter adjustment and heuristic knowledge. Ablation studies are conducted to verify the effectiveness of each component.
Our main contributions are summarized as follows.
\begin{itemize}
\itemsep 0pt
	\item
	We formulate object tracking problem as a classification-only problem, aiming to distinguish the top-left and bottom-right corners of the target from the searching area. The algorithm is general, robust yet simple, achieving better or comparable results to the SOTA algorithms on several mainstream benchmarks. The tracking speed is also impressive with 80 FPS.
    \item
    We design the Pyramid Correlation to equip the correlation feature with fine-grained local structures and global spatial contexts, providing rich information for subsequent processes.
    \item
    We preform Deep Hough Voting on the correlation feature, further capturing the channel-spatial relationships and long range dependencies, enabling the peak position of each feature map to accurately represent the target position.
\end{itemize}

\section{Related work}
\label{related_work}

\paragraph{Siamese-based algorithm}
Recently, Siamese-based trackers have attracted great attention from the visual tracking community due to their satisfactory balance between performance and efficiency. SiamFC~\cite{SiamFC} first learns the similarity map between the target template and the search region through a cross-correlation operation. 
SiamRPN~\cite{Li_2018_CVPR}, SiamRPN++~\cite{Li_2019_CVPR}, DaSiam~\cite{10.1007/978-3-030-01240-3_7}introduce the region proposal extraction subnetwork (RPN)~\cite{7485869} into the Siamese structure, using a classification branch for foreground-background estimation and a regression branch for anchor adjustment. Although these anchor-based algorithms achieve state-of-the-art results on many challenging benchmarks, the pre-defined anchor settings introduce many hyperparameters and computational complexity. Anchor-free tracking algorithms are further raised for better performance. SiamFC++~\cite{Xu_Wang_Li_Yuan_Yu_2020}, SiamCAR~\cite{Guo_2020_CVPR} predict the probability of a point being the target center by classification firstly, and then regress the distances between the center-point and the the object boundaries. The above mentioned algorithms all embed target feature and search feature through cross-correlation, while SiamGAT~\cite{Guo_2021_CVPR} choose to establish part-to-part correspondence between the two Siamese branches feature through graph attention mechanism.

\paragraph{Misalignment of the sibling head}
As is obvious, most of the existing siamese-based tracking algorithms incorporate regression branch, but we hold a different view. 
It turns out through experiments that features in salient areas may have rich classification information, while features around the boundary are more suitable for bounding box regression~\cite{Song_2020_CVPR}.
When the shared features extracted from the Siamese network are applied to both classification and localization branches, the performance is prone to fall into suboptimal due to the essential misalignment of the two branches~\cite{Cheng_2018_ECCV}.
To solve this problem, ~\cite{Song_2020_CVPR} proposes a simple TSD operator to deal with the tangled tasks conflict through a task-aware proposal estimation and a detection head.
~\cite{yan2021learning} trains the network by a two-stage approach through decoupling the classification branch from the regression branch. 
We seek an alternative way to explore a higher performance classifier, using purely classification branch to separate corner points from all pixels. 
The performance could be equally promising attributed to the sophisticated design of the network.

\section{Deep hough voting for visual tracking}
\label{DHV}

\subsection{Overall Architecture}

\begin{figure*}[h]
\begin{center}
    \includegraphics[width=0.85\linewidth]{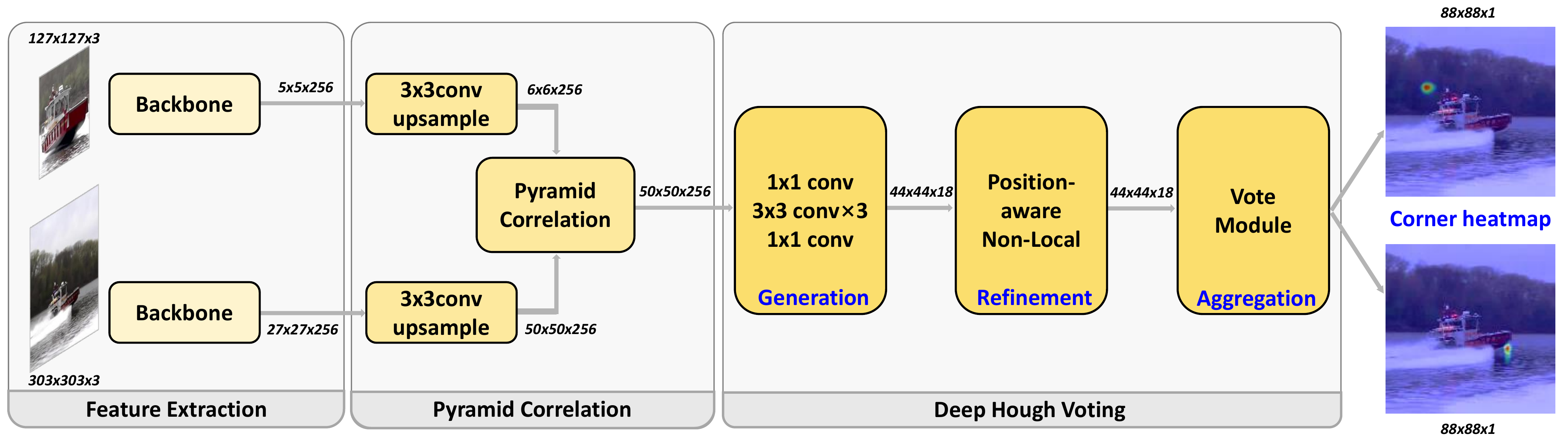}
\end{center}
   \caption{Architecture of our PCDHV tracking framework with three fundamental components: Feature Extraction, Pyramid Correlation and Deep Hough Voting. The peak positions of the two output heatmaps respectively represent the locations of the predicted top-left and bottom-right corners.
   }
\label{fig:overall}
\end{figure*}

Fig.~\ref{fig:overall} shows the overall structure of our algorithm solely composed of a classification branch.
We adopt Siamese structure with parameter-shared backbone network for feature extraction to generate template feature and search feature. An extra CNN-Upsample layer with unshared parameters is used for feature adjustment. 
Then, the Pyramid Correlation module, with blocks of spatial feature selection, pyramid feature pooling and group pixel-level correlation, is applied to obtain corner-favorable correlation feature. 
Finally, the Deep Hough Voting module, with blocks of vote generation, vote refinement and vote aggregation, is applied for accurate corner estimation. 
Two heatmaps respectively corresponding to the top-left corner and bottom-right corner are output for evaluation. 
Without extra tricks, the peak position of each map is assumed to be the location of the predicted corner point.

\subsection{Pyramid correlation}

\begin{figure*}[h]
\begin{center}
\includegraphics[width=1\linewidth]{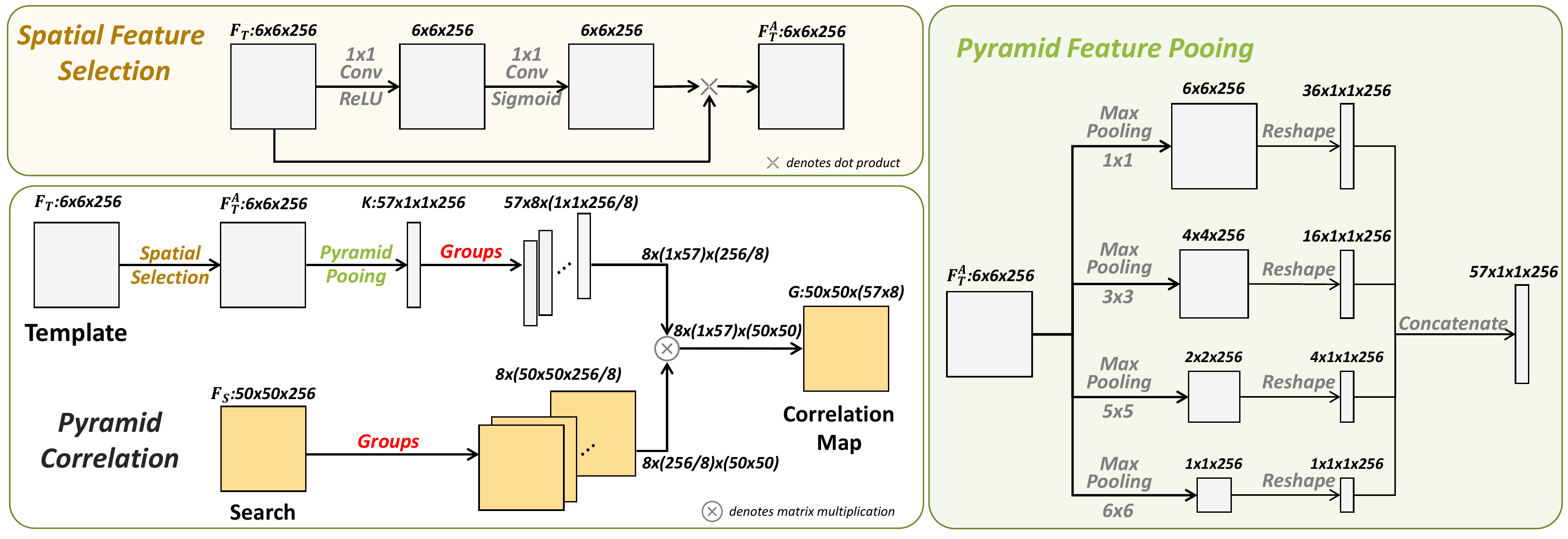}
\end{center}
   \caption{The flowchart of Pyramid Correlation. The template feature is first processed by spatial feature selection, pyramid feature pooling and then correlate with the search feature in a manner of group pixel-level correlation.}
\label{fig:correlation}
\end{figure*}

Fig.~\ref{fig:correlation} depicts the flowchart of our pyramid correlation, which composes of three steps of spatial feature selection, pyramid feature pooling, and group pixel-level correlation. To facilitate notation, 
we denote $\bm{F}_T\in \mathbb{R}^{h \times w \times C}$ as the template feature map and $\bm{F}_S\in \mathbb{R}^{H \times W \times C}$ as the search feature.

\paragraph{Spatial feature selection.}

For precise corner localization, we first highlight the parts of the template that are beneficial to locating corners while suppressing the less useful parts through a spatial attention mechanism.
Concretely, we apply two $1\times1$ convolution, sequentially followed by activation of ReLU and Sigmoid, upon $\bm{F}_T$ to produce a channel-wise spatial attention map, which is used to adjust the importance of the template feature both in a pixel-wise and channel-wise manner. 
As such, the resulting attentive template feature, denoted as $\bm{F}_T^A$, is more prominent and contributes to generating informative correlation maps favoring the identification of target corner points.

\paragraph{Pyramid feature pooling.}

Given $\bm{F}_T^A\in \mathbb{R}^{h \times w \times C}$,
this step aims to construct a set of pixel-level template features comprised of both fine-grained local structures and global spatial contexts.
To this end, we first perform pyramid pooling on $\bm{F}_T^A$, implemented by a group of max-pooling with various odd kernel sizes, leading to a set of pyramid features with decreasing spatial resolutions but enlarging respective fields.
We then decompose these pyramid feature maps in spatial dimension, resulting in a pool of totally $M$ pixel-level feature vectors $\bm{K_i}\in \mathbb{R}^{1 \times 1 \times C}, i\in (0,M-1)$. 
In our implementation, we perform max-pooling with odd kernel sizes of $3\times 3$, $5\times 5$, and global max-pooling with kernel size of $6\times 6$ on $\bm{F}_T^A\in \mathbb{R}^{6 \times 6 \times 256}$, the outputs of which, together with $\bm{F}_T^A$, are then decompose into $57=6^2+4^2+2^2+1$ pixel-level feature vectors for further use.

\paragraph{Group pixel-level correlation.}

This step is designed to perform correlation between the set of pixel-level template features $\bm{K_i}\in \mathbb{R}^{1 \times 1 \times C}, i\in (0,M-1)$ and the 
search feature $\bm{F}_S\in \mathbb{R}^{H \times W \times C}$.
Nevertheless, naive correlation operation produces only one-channel response maps, which causes severe compression or even lost of information. 
To alleviate this, we propose a new group pixel-level correlation by first splitting both the template and search features into $N$ channel groups (we set $N=8$ here), forming $\bm{K_i^N}\in \mathbb{R}^{1 \times 1 \times C/N}$ and $\bm{F}_S^N\in \mathbb{R}^{H \times W\times C/N}$. 
Naive correlation is then performed corresponding in a group-wise manner, producing a $N$-channel correlation map $\bm{G_i}\in \mathbb{R}^{H \times W \times N}$. 
Finally, we concatenate the correlation maps produced by all of the pixel-level template features in the channel dimension, achieving the final correlation results of $MN$ channels, which can be denotes as $\bm{G}\in \mathbb{R}^{H \times W \times MN}$.
Feature scale changes are clearly shown in Fig.~\ref{fig:correlation}.

\subsection{Deep Hough Voting}

The top-left and bottom-right corners can be far away from the target body thus hard to regress directly in a local manner. 
We propose to locate target corners accurately by integrating near and long-range evidences through Hough voting. 
Specifically, the occurrence probability of a target's corner at a given point is determined by the sum of votes received from surroundings.
As with the vote-field with $R$ regions designed in HoughNet~\cite{10.1007/978-3-030-58595-2_25}, votes from both near and far distances can be collected simultaneously.
Fig.~\ref{fig:overall} shows the diagram of our deep hough voting module comprised of three blocks that progressively implement vote generation, vote refinement and vote aggregation, as detailed below.

\paragraph{Vote generation.}
This step aims to generate $H_{S} \times W_{S} \times 2R$ voting map $\bm{F}_{V}$
from the $H \times W \times MN$ 
correlation map, where $H_{S}$ and $W_{S}$ are spatial dimensions, $R$ is the number of regions in the vote filed, and $2$ is the number of corners to be voted for.
We use two convolutional layers with $1 \times 1$ filters for channel reduction, and another three unpadded convolutional layers with $3 \times 3$ filters to enlarge respective field. Therefore, the spatial resolution of the feature is reduced from $H \times W $ to $H_{S} \times W_{S}$.

\begin{figure*}[h]
\begin{center}
\includegraphics[width=0.6\linewidth]{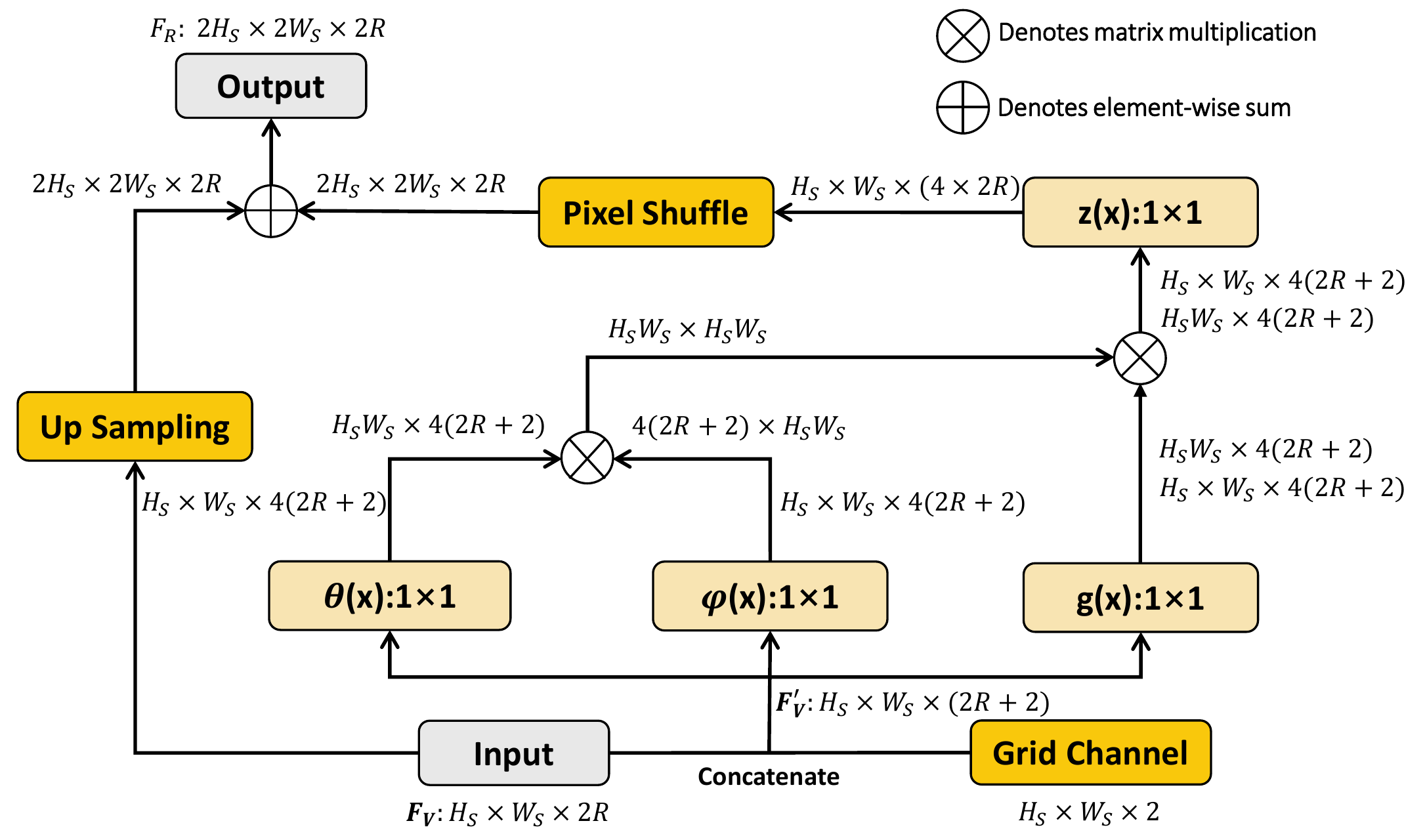}
\end{center}
   \caption{Vote refinement: a modified position-aware non-local block.
   The grid channel and Pixel Shuffle mechanism are deployed respectively to obtain location information and expand feature size.
   }
\label{fig:side:nonlocal}
\end{figure*}

\paragraph{Vote refinement.}
The layers for vote generation are only local operations and may overlook the dependency among votes generated by positions across long distance. Motivated by this, this step refines the votes at all positions in a collaborative manner by capturing their long-range dependencies.
Considering the relations of votes of two positions highly depend on their relative locations, two extra channels are added to $\bm{F}_V$: the normalized $x$ and $y$ coordinates of each spatial position, leading to a $H_{S} \times W_{S} \times  (2R+2)$ sized voting map $\bm{F}_V^{'}$.
We compute interactions between any two positions, and refine the votes at one position as a weighted sum of the votes at all positions, thus obtaining a refined voting map $\bm{F}_R$: 
\begin{equation}
\bm{F}_R = ReLU \{ \mathcal{P}  [ z ( \theta({\bm{F}_V^{'})^T} \phi(\bm{F}_V^{'}) g(\bm{F}_V^{'})) ] + \mathcal{U}[ \bm{F}_V ] \},  
\end{equation}
where $\theta$, $\phi$, $g$ and $z$ represent linear embedding implemented by $1\times1$ convolution.
We set the number of output channels of $z$ to be four times the number of channels in $\bm{F}_V$, and exploit inherent channel-spatial relationships to achieve two-fold spatial expansion of voting map by converting the augmented channel information into space via PixelShuffle operation~\cite{7780576}, as denoted by $\mathcal{P}\left \{\cdot \right \}$.
Accordingly, the residual connection is up-sampled by bi-linear interpolation $\mathcal{U}\left \{\cdot \right \}$.
The summation of the two up-sampled features, activated by ReLU, is taken as output refined voting map, denoted as $\bm{F}_R \in \mathbb{R}^{2H_{S} \times 2W_{S} \times 2R} $. 
\begin{figure*}
\begin{center}
\includegraphics[width=0.65\linewidth]{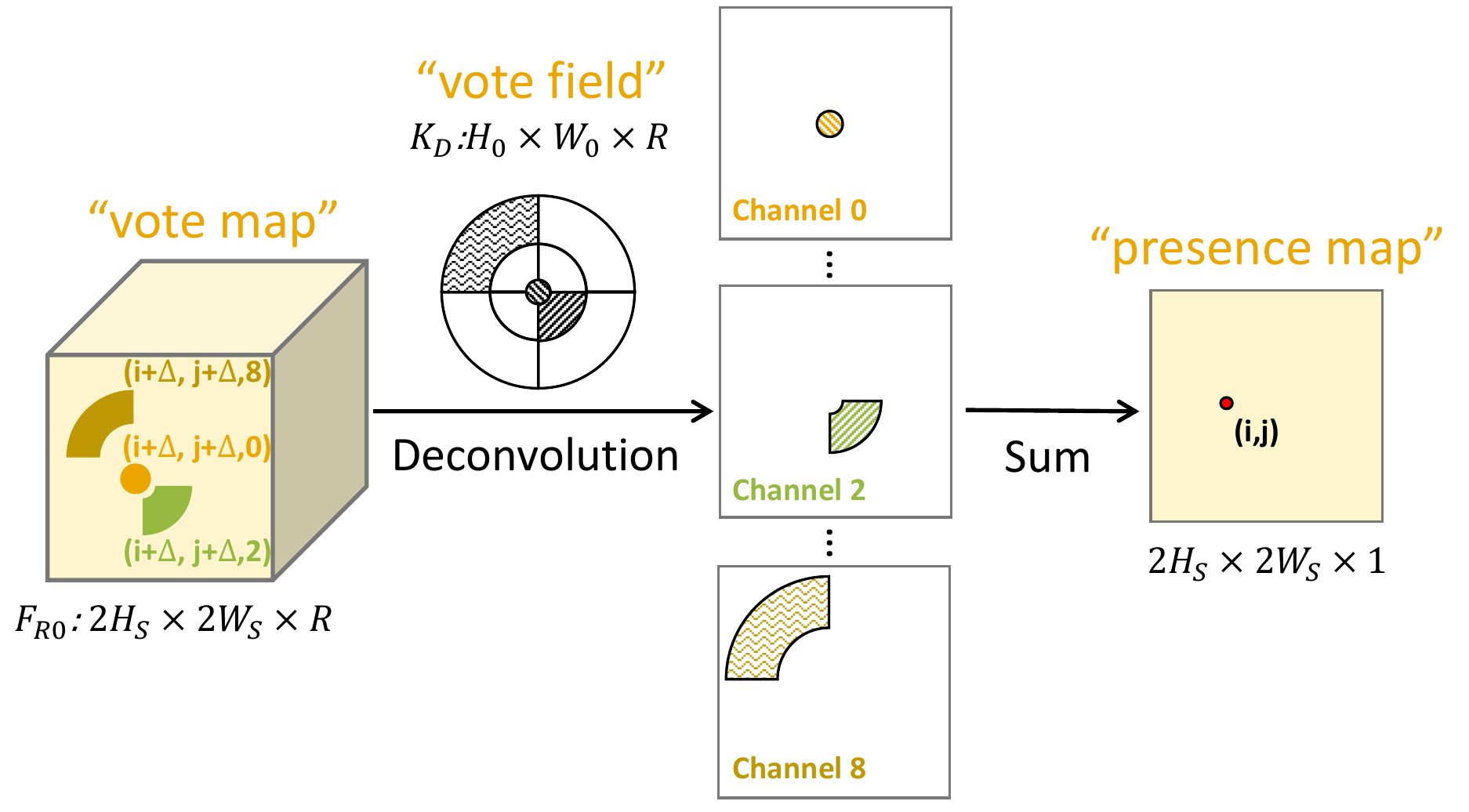}
\end{center}
   \caption{Vote aggregation: the vote module proposed by HoughNet.
   The figure shows an example of the voting process of position $(i,j)$ in top-left corner \emph{presence maps}, where the vote module collects vote from surrounding pixels through a deconvolution operation.
   }
\label{fig:side:houghvote}
\end{figure*}

\paragraph{Vote aggregation.}
This step converts the refined voting map to $2H_{S} \times 2W_{S} \times 1 $  sized \emph{presence maps} for the top-left and bottom-right corner, respectively. Peaks in these maps will indicate the position of target corners.
We take the vote module proposed by HoughNet~\cite{10.1007/978-3-030-58595-2_25} as our vote aggregation block, 
where the set of votes from surrounding pixels can be efficiently aggregated through a deconvolution operation with a fixed-weight (unlearnable) log-polar \emph{vote field} as kernel.
Fig.~\ref{fig:side:houghvote} detailedly illustrates the voting process of position $(i,j)$ in top-left corner \emph{presence maps}, taking $\bm{F}_{R0} \in \mathbb{R}^{2H_{S} \times 2W_{S} \times R} $ (the first R channels of $\bm{F}_{R}$) as input.
To be specific, we slightly adjust the \emph{vote field's} number of ring radii and angle bins in HoughNet~\cite{10.1007/978-3-030-58595-2_25} to form $\bm{K}_{D} \in \mathbb{R}^{H_{0} \times W_{0} \times R} $ and $R=9$ here.
The texture regions in different channels (in \emph{vote field}, where the channel distribution is not explicitly presented) represent the weights of the same shaped color regions in corresponding channels (in \emph{vote map}), from which the target point $(i,j)$ get votes. 
And the sum of all weighted votes corresponds to the final vote value at point $(i,j)$ on the top-left corner \emph{presence map}. $\bm{F}_{R1} \in \mathbb{R}^{2H_{S} \times 2W_{S} \times R} $ ((the rest R channels of $\bm{F}_{R}$)) will be derived by the same operation to get bottom-right corner \emph{presence maps}.

\subsection{Spatial resolution enhancement}
We use GoogLeNet~\cite{2014Going} pretrained on ImageNet~\cite{krizhevsky2012imagenet} as backbone whose stride equals 8. 
The large stride, which results in quantification error when mutually mapping the feature maps to original sequence, may severely hurt subsequent localization accuracy of corners.
Instead of introducing regression branch to bridge this discretization gap, we use a simple yet effective compensation solution by inserting two upsampling operations in the subsequent modules to improve feature resolution, one of which is added right after the backbone feature extraction, and the other is fused into the deep hough voting module as mentioned above.
Finally the total stride is reduced to 2, achieving a balance between accuracy and efficient training.

\subsection{Optimization}
The groundtruth map $Y\in \mathbb{R}^{2H_{S} \times 2W_{S} \times 2}$ is constructed under the instruction of CornerNet~\cite{Law_2018_ECCV}. We first map the corner point coordinates $c(x,y)$ from the search map to the feature map through 
$\lfloor \frac{c(x,y) - \Delta_o}{s} \rfloor$ , where $s$ and $\Delta_o$ respectively denote total stride and total offset of the entire un-padded network, and $\lfloor \cdot \rfloor$ denotes the floor function.
We set the positive area of groundtruth map be a 2D Gaussian kernel of R radius placed at the center of the corner points for efficient training, where R is determined by the at least d IOU between a pair of corners and the target groundtruth (we set $d=0.5$ here). While the rest of the area is considered to be negative.
As with many keypoint-based algorithms, we adopt Focal Loss~\cite{Lin_2017_ICCV} as the training objective.

\section{Experiments}
\label{experiments}

To extensively evaluate the proposed method, we compare our PCDHV with several state-of-the-art trackers on three large-scale datasets of TrackingNet, GOT-10k and LaSOT. 

\subsection{Implementation Details}

\paragraph{Training.}
The network is trained on three currently prevalent large scale video datasets of GOT-10k~\cite{huang2019got}, LaSOT~\cite{fan2019lasot} and TrackingNet~\cite{2018TrackingNet} in form of template-search image pair, 
the deviation between which is restrict within an interval of less than 100 frames.
The input size of template image is $127\times127$ pixels, while the search image is $303\times303$ pixels to capture more corner-related information. 
We adopt GoogLeNet~\cite{2014Going} pre-trained on ImageNet~\cite{krizhevsky2012imagenet} as backbone with a total stride of 8.
Due to the presence of two up sampling operations in the structure, the total stride is reduced to 2, enabling the output feature resolution reach to 88 pixels, which alleviates the impact of discrete sampling while smooths the training process.
As for the training process, convolutional layers in addition to backbone are initialized using a zero-centered Gaussian distribution with a standard deviation of 0.01. 
The network is set to trained 40 epochs in total, each of which contains 150k image pairs. 
The first 5 epochs act as warm up with learning rate linearly increased from $10^{-6}$ to $8 \times 10^{-3}$, while the rest using a cosine annealing learning rate schedule. 
Backbone parameters are unfrozen part by part and eventually all parameters can be updated jointly.
We apply stochastic gradient descent (SGD) with a momentum of 0.9 and a weight decay of $10^{-4}$ for optimization. 
The whole training process is performed on four NVIDIA RTX 2080Ti GPUs.

\paragraph{Testing.}
In the inference phase, the network generates two score maps corresponding to the existence probability of the top-left corner and the bottom-right corner, respectively.
Without bells and whistles, the exact position with the highest score of each is chosen as the position of corner, independent of any form of penalty strategy.
As for the update, a linear interpolation operation is applied to target size to smooth out the variation of the bounding box. Our tracker achieves a competitive speed of over 80 FPS.

\subsection{Evaluation on TrackingNet, GOT-10k and LaSOT Datasets}

\paragraph{Results on TrackingNet.}

TrackingNet~\cite{2018TrackingNet} contains 30,000 sequences with 14 million dense annotations and a test set of 511 sequences. It covers different object classes and scenes and requires the tracker to have both discriminative and generative capabilities. Precision, normalized precision, and AUC scores are used to evaluate trackers. 
The results on TrackingNet are shown in Table \ref{tab:trackingnet}.
It can be seen that we outperform our competitor CGACD~\cite{Du_2020_CVPR}, a corner-based tracking algorithm as well, by a large margin of 4.5\%. 
Moreover, compared with the state-of-the-art Siamese based tracking algorithms of SiamAttn~\cite{Yu_2020_CVPR} and SiamGAT\cite{Guo_2021_CVPR}, we performs the best in both $AUC$ and $P$, but regrettably, less well in $P_{norm}$. 
We argue the reason is that $P_{norm}$ appreciates models friendly to targets with large scale variations.
Since our work pursues solutions with high simplicity and generality, we directly apply vanilla backbones for feature extraction without any additional tailored components, despite not ranking first in every metric, but achieving the ideal trade-off between accuracy and efficiency.
The leading results on such a large dataset also illustrate the well generalization ability of our tracking algorithm.

\begin{table*}[t]
\centering
\footnotesize
\setlength{\tabcolsep}{1mm}
\caption{Performance comparisons on TrackingNet test set. The top three results are highlighted in \textcolor{red}{red}, \textcolor{green}{green} and \textcolor{blue}{blue}, respectively.}
\begin{adjustbox}{max width=1.0\textwidth}
\begin{tabular}{c|ccccccccccc}
\hline
\cline{1-12}
              & \makecell[c]{SiamFC} 
              & \makecell[c]{SiamRPN++} 
              & \makecell[c]{ATOM } 
              & \makecell[c]{DiMP}
              & \makecell[c]{SiamFC++} 
              & \makecell[c]{D3S} 
              & \makecell[c]{CGACD} 
              & \makecell[c]{KYS }
              & \makecell[c]{SiamAttn*}
              & \makecell[c]{SiamGAT*}
              & PCDHV \\   
\hline
\cline{1-12}
AUC   & 57.1    & 73.3   & 70.3   & 74.0   & \textcolor{green}{75.4}  & 72.8  & 71.1 
& 74.0   & 74.3 & \textcolor{blue}{75.3}   & \textcolor{red}{76.0} \\
Pnorm & 66.3   & 80.0   & 77.1   & 80.1   & 80.0  & 76.8  & 80.0   & 80.0  &\textcolor{red}{81.0} & \textcolor{green}{80.7} & \textcolor{blue}{80.4} \\
P     & 53.3   & 69.4   & 64.8   & 68.7   & \textcolor{green}{70.5}  & 66.4  & 69.3   & 68.8 & \textcolor{green}{70.5} & \textcolor{blue}{69.8} & \textcolor{red}{72.1} \\    
\hline
\cline{1-12}

\end{tabular}
\end{adjustbox}

\begin{tablenotes}   
        \scriptsize               
        \item * The performances of SiamAttn and SiamGAT are the results of our reproduction on the TrackingNet test set using the GOT-10k Python toolkit, same as PCDHV.
\end{tablenotes} 

\label{tab:trackingnet}
\end{table*}

\paragraph{Results on GOT-10k.}

GOT-10k~\cite{huang2019got} is a challenging large-scale dataset which contains 10,000 videos in train subset and 180 videos in both val and test subset, all of which are moving objects in real-world.
Since there is no class intersection between its train and test subsets, which contributes to its difficulty, the tracking result can reflect the generalization ability of the algorithm to unseen object classes.
Average Overlap ($AO$) and Success Rate ($SR$) are adopted as performance metrics, with higher values resulting in better performance. 
We follow the protocol of GOT-10k and train our model with only its training subset.
The results on GOT-10k are shown in Table \ref{tab:got10k}.
It can be seen that our PCDHV, though slightly inferior in $AO$ and $SR50$, can perform the best in $SR75$. 
We argue that trackers with individual mechanism of distinguish foreground from background may enjoy outstanding performance on recalls ($AO$ and $SR50$).
While our straightforward PCDHV prefers to well respect the edge (corner) information and thus enjoys powerful localization capability, leading to higher $SR75$.
We further test the speed of several preeminent tracking methods using GOT-10k test set on a single RTX-2080Ti GPU. The last line in Table \ref{tab:got10k} shows our PCDHV clearly stands out regarding accuracy and speed trade-off.


\begin{table*}[h]
\centering
\small
\setlength{\tabcolsep}{0.7mm}
\caption{Performance comparisons on GOT-10k test set. The top three results are highlighted in \textcolor{red}{red}, \textcolor{green}{green} and \textcolor{blue}{blue}, respectively. }
\begin{adjustbox}{max width=1.0\textwidth}
\begin{tabular}{c|ccccccccccc}
\hline
\cline{1-12}
              & \makecell[c]{SiamFC} 
              & \makecell[c]{ECO} 
              & \makecell[c]{SiamRPN++} 
              & \makecell[c]{ATOM } 
              & \makecell[c]{SiamCAR}  
              & \makecell[c]{Ocean-offline} 
              & \makecell[c]{SiamFC++ } 
              & \makecell[c]{D3S} 
              & \makecell[c]{SiamAttn*} 
              & \makecell[c]{SiamGAT} 
              & PCDHV  \\
\hline
\cline{1-12}
AO   & 34.8   & 31.6   & 51.7   & 55.6   & 56.9   & 59.2  & 59.5  & \textcolor{blue}{59.7}  & 59.3 & \textcolor{red}{62.7}   & \textcolor{green}{60.9} \\
$SR_{50}$ & 35.3   & 30.9   & 61.6   & 63.4   & 67.0   & 69.5  & 69.5  & 67.6 &  \textcolor{blue}{70.0}  & \textcolor{red}{74.3}  & \textcolor{green}{71.3} \\
$SR_{75}$ & 9.8    & 11.1   & 32.5   & 40.2   & 41.5   & 47.3  & \textcolor{blue}{47.9}  & 46.2  & 45.6 & \textcolor{green}{48.8}  & \textcolor{red}{50.2} \\
\hline
$FPS$ & - & - & 33 & 25 & - & - & 65 & - & 19 & 49 & 80 \\
\hline
\cline{1-12}
\end{tabular}
\end{adjustbox}

\begin{tablenotes}    
        \scriptsize            
        \item * The performance of SiamAttn is the result of our reproduction on the GOT-10k test set using the GOT-10k Python toolkit, same as PCDHV.
\end{tablenotes}

\label{tab:got10k}
\end{table*}

\begin{figure*}[h]
\begin{center}
\includegraphics[width=0.9\linewidth]{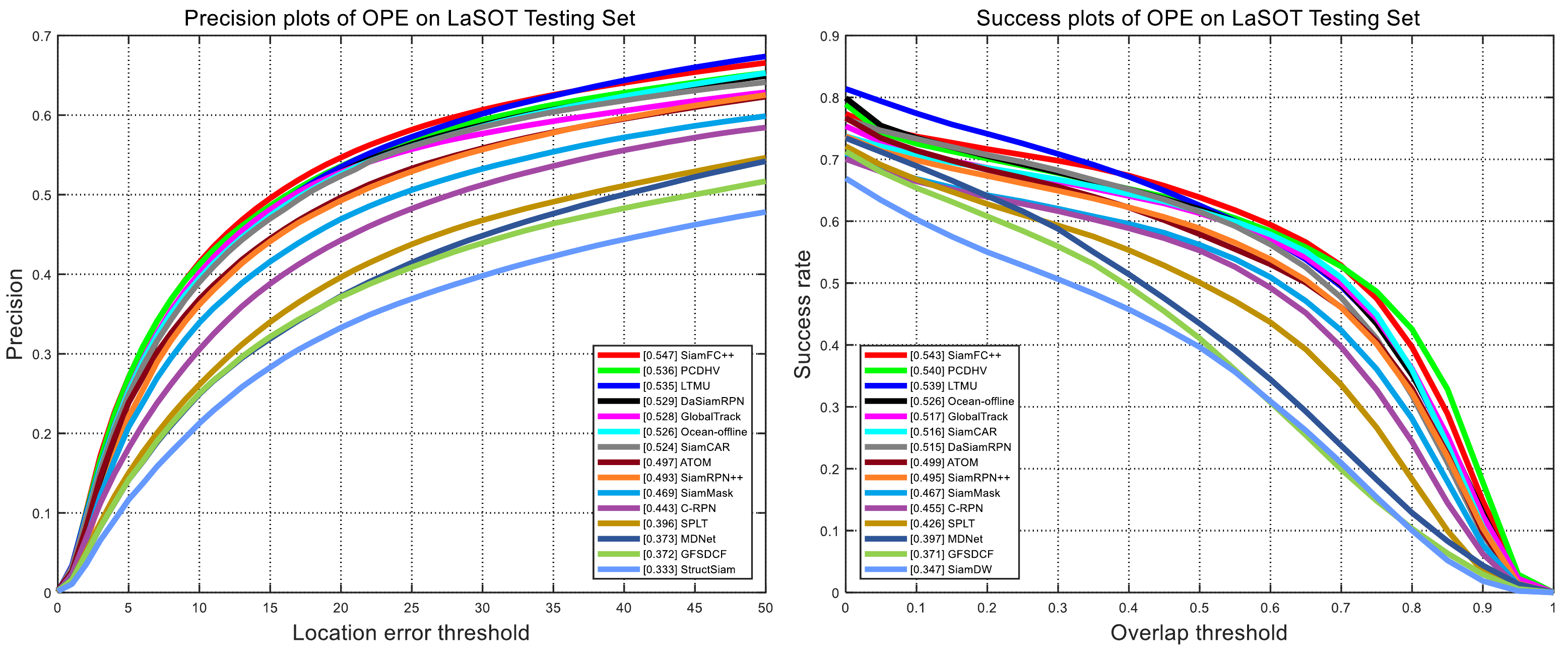}
\end{center}
   \caption{Precision and success plots on LaSOT.}
\label{fig:side:lasot}
\end{figure*}

\paragraph{Results on LaSOT.}
LaSOT~\cite{fan2019lasot} a large-scale, high-quality, and densely annotated dataset for long-term tracking, with 
1,400 sequences under Protocol I while 280 under Protocol II. Each of these sequences is long, with an average of 2,500 frames, facilitating the detection of the tracker's long-term performance. Success and Precision are used for evaluation. 
For fair comparison, we follow Protocol II under which trackers are trained on LaSOT train subset and evaluated on LaSOT test subset. The results on LaSOT are shown in Fig.\ref{fig:side:lasot}.
Compared with SOTA anchor-based tracking algorithm SiamRPN++\cite{Li_2019_CVPR}, our tracker improves the AUC score by 4.4 points,
which is comparable to the results of the SOTA anchor-free tracking algorithm SiamFC++~\cite{ Xu_Wang_Li_Yuan_Yu_2020}.

\paragraph{Analysis and Discussion.}

\begin{figure}[h]
  \begin{minipage}[t]{0.45\linewidth}
    \centering
    \includegraphics[width=1\linewidth]{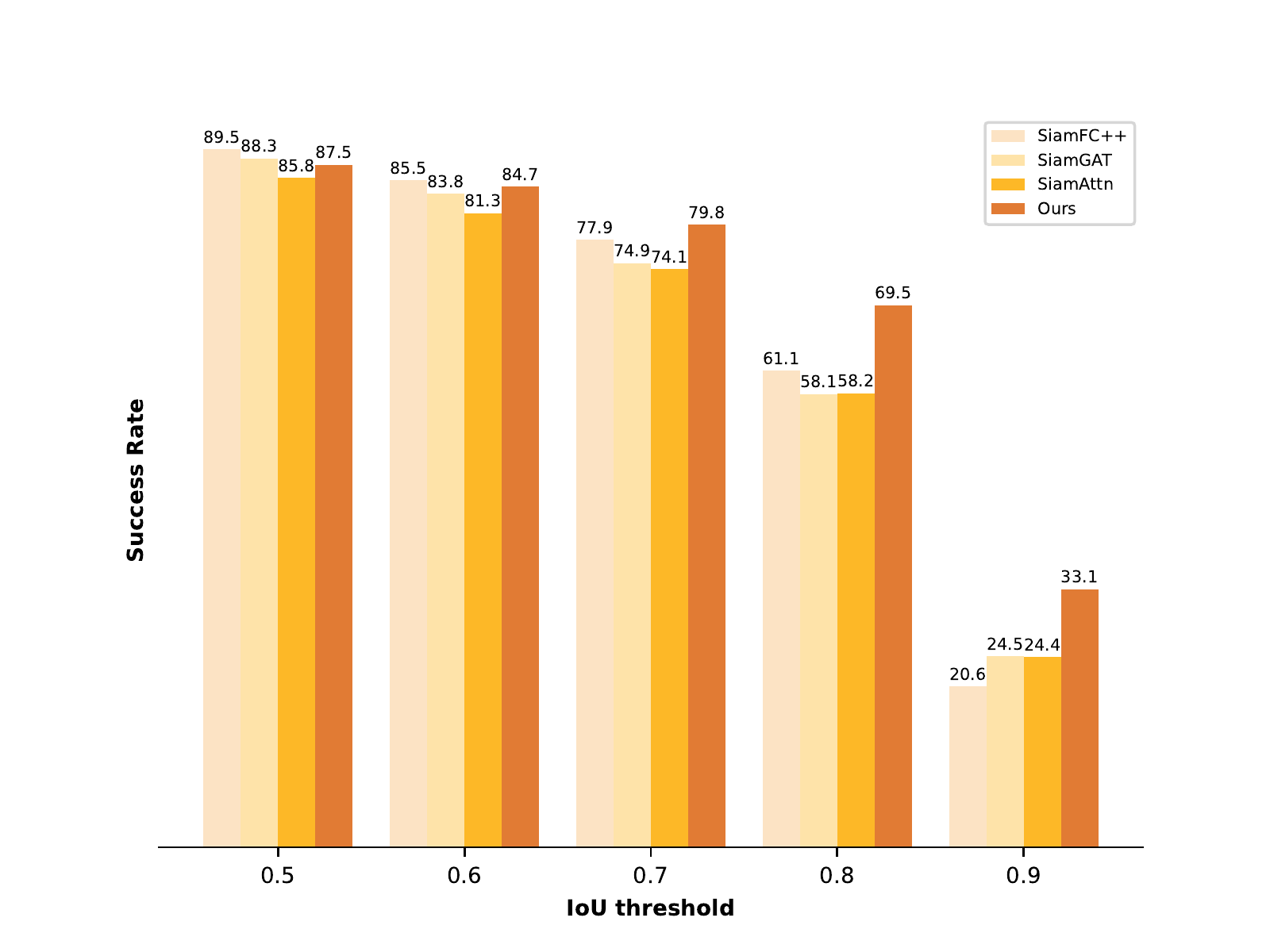}
    \caption{SR across IoU thresholds\\ on GOT-10k val set.}
    \label{fig:side:comparison}
  \end{minipage}%
  \begin{minipage}[t]{0.52\linewidth}
    \centering
    \includegraphics[width=1\linewidth]{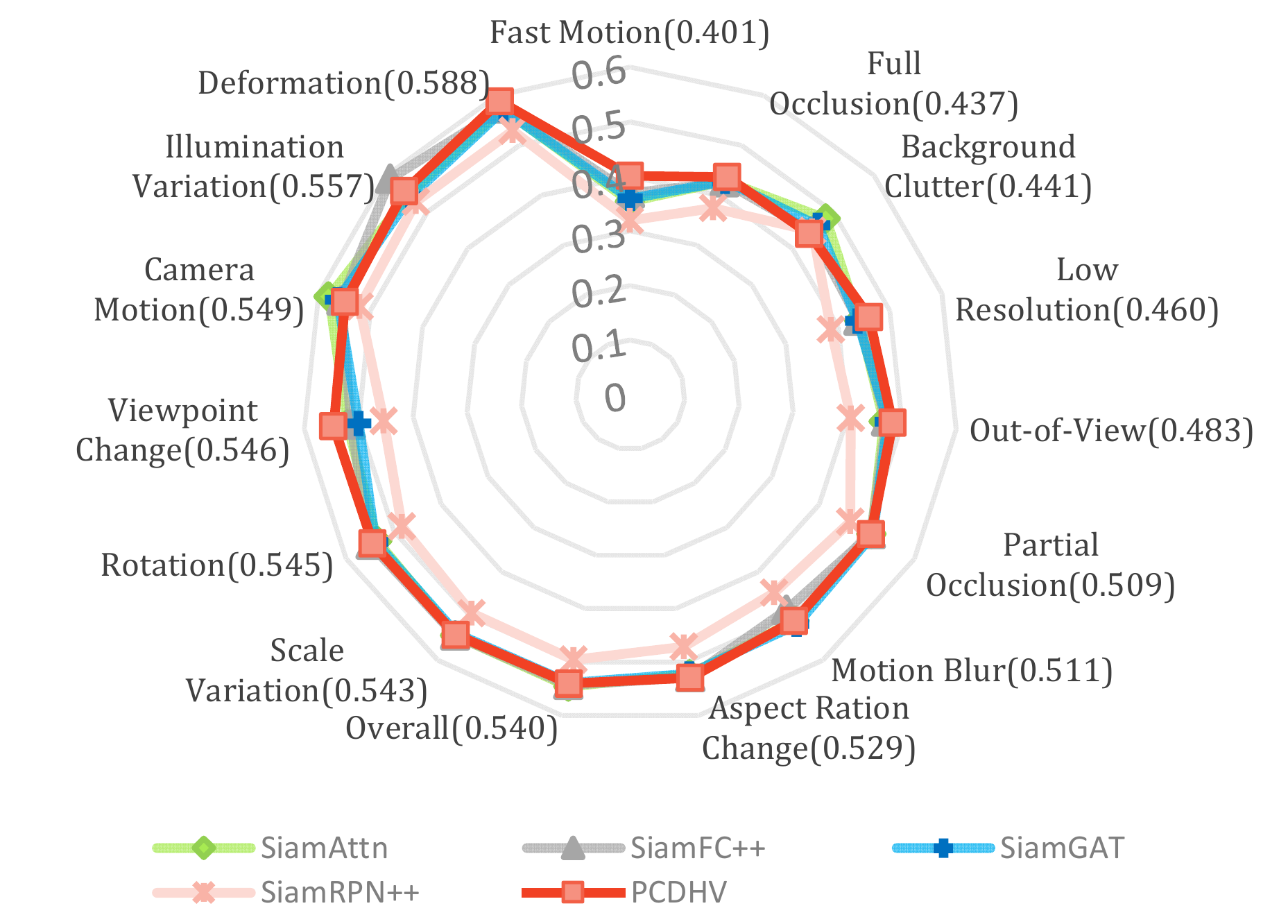}
    \caption{Attribute analysis on LaSOT test set.}
    \label{fig:side:attribute}
  \end{minipage}
\end{figure}

It is worth noting that the performance of our algorithm at higher IoU thresholds is outstanding, which can be seen from metrics such as $SR_{75}$ in GOT-10k (Table \ref{tab:got10k}) and the latter half of the success curves in LaSOT (Fig.\ref{fig:side:lasot}). 
We gather more results of $SR$ on GOT-10k val set, see Fig.\ref{fig:side:comparison}.
The figure shows our PCDHV achieves significant higher rates at large IoU thresholds, which clearly proves its superiority in precisely locating targets.
We argue that this impressive result mainly brought by the robust and informative features and its increased resolution, as can be illustrated by the ablation experiments.
We also perform attribute analysis of our PCDHV and several competing counterparts on LaSOT test set (with the $AUC$ of PCDHV reported following every attribute name), as shown in Fig.\ref{fig:side:attribute}.
Our PCDHV performs better than other prior arts on most attributes.
While in the case of \textit{fast motion}, \textit{full occlusion} and \textit{background clustering}, all trackers behave relatively poorly.
We assume them as common challenges faced by offline tracking algorithms without global search or re-detection mechanism.

\subsection{Ablation Studies}

We perform a component-wise analysis on the GOT-10k benchmark~\cite{huang2019got} which can only be evaluated using an online server, enhancing the fairness and credibility of the test results.
Experiments are conducted through gradually adding each component to the baseline model to independently verify the effectiveness of our Pyramid Correlation and Deep Hough Voting.

\paragraph{Pyramid correlation.}
The baseline model (line 1 in Table \ref{tab:ablation-corr-combine}) here is obtained by degenerating the pyramid correlation into a depth-wise correlation~\cite{Li_2019_CVPR} while preserving other structures. As listed in Table \ref{tab:ablation-corr-combine}, the tracker gained 1.3\%, 1.2\%, 0.6\%, and 1.4\% improvement in $AO$ with the addition operations of pixel-wise correlation, pyramid pooling, attention mechanism and grouping operation, respectively. 
Each block is thus proved to have contribution on the final result through increases the fine-grained local details and the global spatial context on features.

\begin{table*}[h]
\centering
\small
\setlength{\tabcolsep}{1.0mm}
\caption{Effectiveness of each component of Pyramid correlation. }
\begin{adjustbox}{max width=1.0\textwidth}
\begin{tabular}{c|cccc|cc}
\toprule
&Pixel-Wise & Pyramid& Attention & Group & AO & $ \rm SR_{75}$ \\
\midrule
 \multirow{5}{*}{\textbf{Baseline}} &  &  &  & & 56.7  & 46.7 \\
&\checkmark  &  &  &  & 58.0  & 48.1  \\
&\checkmark & \checkmark &  &  & 59.2  & 48.7 \\ 
&\checkmark & \checkmark & \checkmark &  & 59.5  & 50.0 \\  
&\checkmark & \checkmark & \checkmark & \checkmark & \textbf{60.9} & \textbf{50.2} \\
\bottomrule
\end{tabular}
\end{adjustbox}
\label{tab:ablation-corr-combine}
\end{table*}

\noindent
\paragraph{Deep Hough Voting.}
As the vote aggregation block essentially has a quantitative requirement of $2R$ on the channel of the input feature, we degrade the vote generation block into a rough $1\times1$ convolutional layer merely responsible for channel reduction, combine which with the vote aggregation block as the baseline model.
It can be seen from line 2 in Table \ref{tab:ablation-vote-combine} that, despite the feature size shrinkage brought by the un-padded $3\times3$ convolutional layers, our vote generation block can still work well with vote aggregation module and achieves a performance lift of 0.7\% on $AO$, which indicates that the gain in perceptual field increment is greater than the gain in feature size increment. 
An notable improvement achieved through the incorporation of position-aware non-local block in vote refinement block, as shown in line 4, which can be contributed by both operations of grid fusing (3.2\% gain on $AO$, as shown in line 3) and up sampling (2.6\% gain on $AO$, comparing line 3 with line 4).

\begin{table*}[h]
\centering
\small
\setlength{\tabcolsep}{1.0mm}
\caption{Effectiveness of each component of Deep Hough Voting.}
\begin{adjustbox}{max width=1.0\textwidth}
\begin{threeparttable}
\begin{tabular}{c|ccc|cc}
    \toprule
    &Generation     & Refinemen     & Aggregation & AO & $ \rm SR_{75}$\\
    \midrule
     \multirow{4}{*}{\textbf{Baseline}} &depth-wise &  & \checkmark & 54.4  & 43.1    \\
    &\checkmark &  & \checkmark & 55.1 &  45.5     \\
    &\checkmark  &\checkmark(-up)\tnote{*}  &\checkmark  &58.3  &48.1      \\
    &\checkmark & \checkmark & \checkmark & \textbf{60.9}  & \textbf{50.2} \\
    
    \bottomrule
\end{tabular}

\begin{tablenotes}    
        \scriptsize              
        \item[*] \emph{
                      \checkmark(-up) denotes remove upsample from refinement block } 
      \end{tablenotes} 
      
\end{threeparttable}

\end{adjustbox}

\label{tab:ablation-vote-combine}
\end{table*}

\subsection{Qualitative Results}

\begin{figure*}[h]
\begin{center}
\includegraphics[width=0.9\linewidth]{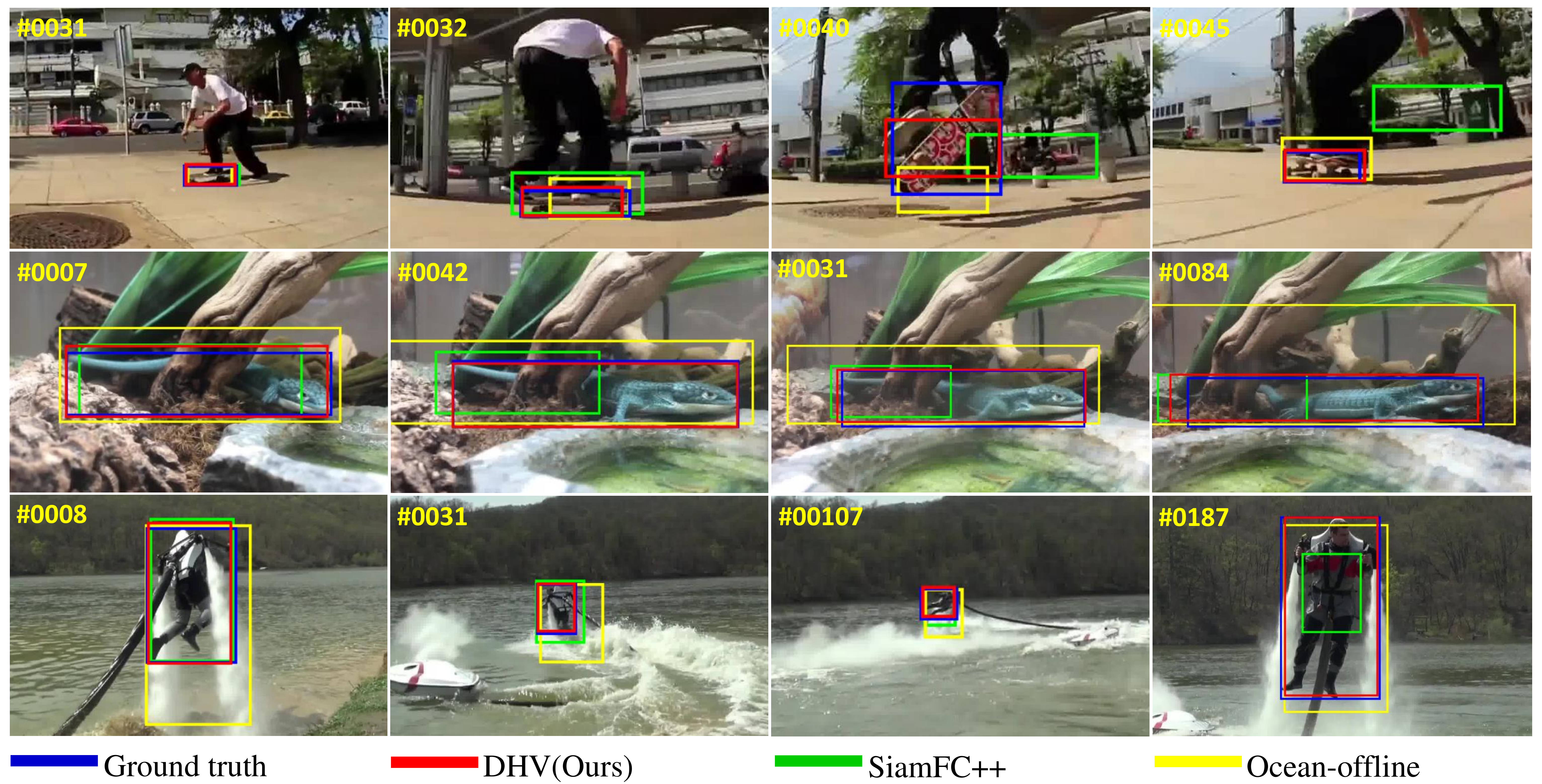}
\end{center}
  \caption{Qualitative results compare with Ocean and SiamFC++ on three challenging sequences in GOT-10k val set.}
\label{fig:qualitative_result}
\end{figure*}

Fig.~\ref{fig:qualitative_result} qualitatively shows our tracking results compared with state-of-the-art trackers Ocean~\cite{10.1007/978-3-030-58589-1_46} and SiamFC++~\cite{Xu_Wang_Li_Yuan_Yu_2020} on three challenging sequences in GOT-10k val set~\cite{huang2019got}. 
It is challenging for offline trackers to accurately assess the exact position of the target when it undergoes large deformation (first row), occlusion (second row) and out-of-plane (third row) changes.
Our tracker performs well than the trendy and advanced anchor-free algorithm, while maintaining a very accurate target position estimation and impressive tracking speed.

\section{Conclusion}
In this paper, we introduce a novel vote-based classification-only tracking algorithm PCDHV, locating the target by accurately estimating their top-left and bottom-right corner. We innovatively proposal a Pyramid Correlation to equip the correlation feature with fine-grained local structure and global spatial context. Then, the Deep Hough Voting takes over and further captures the channel-spatial relationships and long ranger dependencies, realizing that the maximum value of the output feature can accurately correspond to the target location.
The discretization gap between feature maps and original tracking sequence is mitigated by two well-designed up sampling mechanisms. Extensive experiments show that PCDHV achieves better or comparable results than SOTA algorithm on several mainstream datasets.
The tracking speed is also impressive with 80 FPS. 
We will try to further explore the adaptability of deeper feature extraction networks to our algorithm with an expectation of better performance improvement.

\acks{
This work was supported by the Key Laboratory Foundation under Grant TCGZ2020C004, 202020429036, 3040012222101 and 3040036722103.
}

\bibliography{acml21}

\begin{thebibliography}{30}
\providecommand{\natexlab}[1]{#1}
\providecommand{\url}[1]{\texttt{#1}}
\expandafter\ifx\csname urlstyle\endcsname\relax
  \providecommand{\doi}[1]{doi: #1}\else
  \providecommand{\doi}{doi: \begingroup \urlstyle{rm}\Url}\fi

\bibitem[Bertinetto et~al.(2016)Bertinetto, Valmadre, Henriques, Vedaldi, and
  Torr]{SiamFC}
Luca Bertinetto, Jack Valmadre, Jo{\~a}o~F Henriques, Andrea Vedaldi, and
  Philip H~S Torr.
\newblock Fully-convolutional siamese networks for object tracking.
\newblock In \emph{ECCV}, 2016.

\bibitem[Cheng et~al.(2018)Cheng, Wei, Shi, Feris, Xiong, and
  Huang]{Cheng_2018_ECCV}
Bowen Cheng, Yunchao Wei, Honghui Shi, Rogerio Feris, Jinjun Xiong, and Thomas
  Huang.
\newblock Revisiting rcnn: On awakening the classification power of faster
  rcnn.
\newblock In \emph{ECCV}, 2018.

\bibitem[Danelljan et~al.(2019)Danelljan, Bhat, Khan, and
  Felsberg]{Danelljan_2019_CVPR}
Martin Danelljan, Goutam Bhat, Fahad~Shahbaz Khan, and Michael Felsberg.
\newblock Atom: Accurate tracking by overlap maximization.
\newblock In \emph{CVPR}, 2019.

\bibitem[Du et~al.(2020)Du, Liu, Zhao, and Tang]{Du_2020_CVPR}
Fei Du, Peng Liu, Wei Zhao, and Xianglong Tang.
\newblock Correlation-guided attention for corner detection based visual
  tracking.
\newblock In \emph{CVPR}, 2020.

\bibitem[Fan et~al.(2019)Fan, Lin, Yang, Chu, Deng, Yu, Bai, Xu, Liao, and
  Ling]{fan2019lasot}
Heng Fan, Liting Lin, Fan Yang, Peng Chu, Ge~Deng, Sijia Yu, Hexin Bai, Yong
  Xu, Chunyuan Liao, and Haibin Ling.
\newblock Lasot: A high-quality benchmark for large-scale single object
  tracking.
\newblock In \emph{CVPR}, 2019.

\bibitem[Guo et~al.(2020)Guo, Wang, Cui, Wang, and Chen]{Guo_2020_CVPR}
Dongyan Guo, Jun Wang, Ying Cui, Zhenhua Wang, and Shengyong Chen.
\newblock Siamcar: Siamese fully convolutional classification and regression
  for visual tracking.
\newblock In \emph{CVPR}, 2020.

\bibitem[Guo et~al.(2021)Guo, Shao, Cui, Wang, Zhang, and Shen]{Guo_2021_CVPR}
Dongyan Guo, Yanyan Shao, Ying Cui, Zhenhua Wang, Liyan Zhang, and Chunhua
  Shen.
\newblock Graph attention tracking.
\newblock In \emph{CVPR}, 2021.

\bibitem[Gupta et~al.(2017)Gupta, Kumar, Behera, and Subramanian]{7637024}
Meenakshi Gupta, Swagat Kumar, Laxmidhar Behera, and Venkatesh~K. Subramanian.
\newblock A novel vision-based tracking algorithm for a human-following mobile
  robot.
\newblock \emph{IEEE Transactions on Systems, Man, and Cybernetics: Systems},
  47\penalty0 (7):\penalty0 1415--1427, 2017.

\bibitem[Huang et~al.(2019)Huang, Zhao, and Huang]{huang2019got}
Lianghua Huang, Xin Zhao, and Kaiqi Huang.
\newblock Got-10k: A large high-diversity benchmark for generic object tracking
  in the wild.
\newblock \emph{IEEE Transactions on Pattern Analysis and Machine
  Intelligence}, 2019.

\bibitem[Kiani~Galoogahi et~al.(2017)Kiani~Galoogahi, Fagg, and
  Lucey]{kiani2017learning}
Hamed Kiani~Galoogahi, Ashton Fagg, and Simon Lucey.
\newblock Learning background-aware correlation filters for visual tracking.
\newblock In \emph{ICCV}, 2017.

\bibitem[Krizhevsky et~al.(2012)Krizhevsky, Sutskever, and
  Hinton]{krizhevsky2012imagenet}
Alex Krizhevsky, Ilya Sutskever, and Geoffrey~E Hinton.
\newblock Imagenet classification with deep convolutional neural networks.
\newblock \emph{Advances in neural information processing systems},
  25:\penalty0 1097--1105, 2012.

\bibitem[Law and Deng(2018)]{Law_2018_ECCV}
Hei Law and Jia Deng.
\newblock Cornernet: Detecting objects as paired keypoints.
\newblock In \emph{ECCV}, 2018.

\bibitem[Li et~al.(2018)Li, Yan, Wu, Zhu, and Hu]{Li_2018_CVPR}
Bo~Li, Junjie Yan, Wei Wu, Zheng Zhu, and Xiaolin Hu.
\newblock High performance visual tracking with siamese region proposal
  network.
\newblock In \emph{CVPR}, 2018.

\bibitem[Li et~al.(2019)Li, Wu, Wang, Zhang, Xing, and Yan]{Li_2019_CVPR}
Bo~Li, Wei Wu, Qiang Wang, Fangyi Zhang, Junliang Xing, and Junjie Yan.
\newblock Siamrpn++: Evolution of siamese visual tracking with very deep
  networks.
\newblock In \emph{CVPR}, 2019.

\bibitem[Li et~al.(2013)Li, Hu, Shen, Zhang, Dick, and Hengel]{li2013survey}
Xi~Li, Weiming Hu, Chunhua Shen, Zhongfei Zhang, Anthony Dick, and Anton
  Van~Den Hengel.
\newblock A survey of appearance models in visual object tracking.
\newblock \emph{ACM transactions on Intelligent Systems and Technology (TIST)},
  4\penalty0 (4):\penalty0 1--48, 2013.

\bibitem[Lin et~al.(2017)Lin, Goyal, Girshick, He, and Dollar]{Lin_2017_ICCV}
Tsung-Yi Lin, Priya Goyal, Ross Girshick, Kaiming He, and Piotr Dollar.
\newblock Focal loss for dense object detection.
\newblock In \emph{ICCV}, 2017.

\bibitem[Mueller et~al.(2018)Mueller, Bibi, Giancola, Alsubaihi, and
  Ghanem]{2018TrackingNet}
Matthias Mueller, A.~Bibi, S.~Giancola, S.~Alsubaihi, and B.~Ghanem.
\newblock Trackingnet: A large-scale dataset and benchmark for object tracking
  in the wild.
\newblock In \emph{ECCV}, 2018.

\bibitem[Ren et~al.(2017)Ren, He, Girshick, and Sun]{7485869}
Shaoqing Ren, Kaiming He, Ross Girshick, and Jian Sun.
\newblock Faster r-cnn: Towards real-time object detection with region proposal
  networks.
\newblock \emph{IEEE Transactions on Pattern Analysis and Machine
  Intelligence}, 39\penalty0 (6):\penalty0 1137--1149, 2017.

\bibitem[Samet et~al.(2020)Samet, Hicsonmez, and
  Akbas]{10.1007/978-3-030-58595-2_25}
Nermin Samet, Samet Hicsonmez, and Emre Akbas.
\newblock Houghnet: Integrating near and long-range evidence for bottom-up
  object detection.
\newblock In Andrea Vedaldi, Horst Bischof, Thomas Brox, and Jan-Michael Frahm,
  editors, \emph{ECCV}, 2020.

\bibitem[Shi et~al.(2016)Shi, Caballero, Huszár, Totz, Aitken, Bishop,
  Rueckert, and Wang]{7780576}
Wenzhe Shi, Jose Caballero, Ferenc Huszár, Johannes Totz, Andrew~P. Aitken,
  Rob Bishop, Daniel Rueckert, and Zehan Wang.
\newblock Real-time single image and video super-resolution using an efficient
  sub-pixel convolutional neural network.
\newblock In \emph{CVPR}, 2016.

\bibitem[Song et~al.(2020)Song, Liu, and Wang]{Song_2020_CVPR}
Guanglu Song, Yu~Liu, and Xiaogang Wang.
\newblock Revisiting the sibling head in object detector.
\newblock In \emph{CVPR}, 2020.

\bibitem[Szegedy et~al.(2014)Szegedy, Liu, Jia, Sermanet, and
  Rabinovich]{2014Going}
Christian Szegedy, Wei Liu, Yangqing Jia, Pierre Sermanet, and Andrew
  Rabinovich.
\newblock Going deeper with convolutions.
\newblock \emph{IEEE Computer Society}, 2014.

\bibitem[Voigtlaender et~al.(2020)Voigtlaender, Luiten, Torr, and
  Leibe]{Voigtlaender_2020_CVPR}
Paul Voigtlaender, Jonathon Luiten, Philip~H.S. Torr, and Bastian Leibe.
\newblock Siam r-cnn: Visual tracking by re-detection.
\newblock In \emph{CVPR}, 2020.

\bibitem[Wu et~al.(2015)Wu, Lim, and Yang]{7001050}
Yi~Wu, Jongwoo Lim, and Ming-Hsuan Yang.
\newblock Object tracking benchmark.
\newblock \emph{IEEE Transactions on Pattern Analysis and Machine
  Intelligence}, 37\penalty0 (9):\penalty0 1834--1848, 2015.

\bibitem[Xu et~al.(2020)Xu, Wang, Li, Yuan, and Yu]{Xu_Wang_Li_Yuan_Yu_2020}
Yinda Xu, Zeyu Wang, Zuoxin Li, Ye~Yuan, and Gang Yu.
\newblock Siamfc++: Towards robust and accurate visual tracking with target
  estimation guidelines.
\newblock In \emph{AAAI}, 2020.

\bibitem[Yan et~al.(2021)Yan, Peng, Fu, Wang, and Lu]{yan2021learning}
Bin Yan, Houwen Peng, Jianlong Fu, Dong Wang, and Huchuan Lu.
\newblock Learning spatio-temporal transformer for visual tracking, 2021.

\bibitem[Yu et~al.(2020)Yu, Xiong, Huang, and Scott]{Yu_2020_CVPR}
Yuechen Yu, Yilei Xiong, Weilin Huang, and Matthew~R. Scott.
\newblock Deformable siamese attention networks for visual object tracking.
\newblock In \emph{CVPR}, 2020.

\bibitem[Zhang et~al.(2014)Zhang, Zhang, and Yang]{6784124}
Kaihua Zhang, Lei Zhang, and Ming-Hsuan Yang.
\newblock Fast compressive tracking.
\newblock \emph{IEEE Transactions on Pattern Analysis and Machine
  Intelligence}, 36\penalty0 (10):\penalty0 2002--2015, 2014.

\bibitem[Zhang et~al.(2020)Zhang, Peng, Fu, Li, and
  Hu]{10.1007/978-3-030-58589-1_46}
Zhipeng Zhang, Houwen Peng, Jianlong Fu, Bing Li, and Weiming Hu.
\newblock Ocean: Object-aware anchor-free tracking.
\newblock In \emph{ECCV}, 2020.

\bibitem[Zhu et~al.(2018)Zhu, Wang, Li, Wu, Yan, and
  Hu]{10.1007/978-3-030-01240-3_7}
Zheng Zhu, Qiang Wang, Bo~Li, Wei Wu, Junjie Yan, and Weiming Hu.
\newblock Distractor-aware siamese networks for visual object tracking.
\newblock In \emph{ECCV}, 2018.

\end{thebibliography}






\end{document}